# Edge-detection applied to moving sand dunes on Mars

**Amelia Carolina Sparavigna**

Department of Applied Science and Technology, Politecnico di Torino, Torino, Italy

**Abstract:** Here we discuss the application of an edge detection filter, the Sobel filter of GIMP, to the recently discovered motion of some sand dunes on Mars. The filter allows a good comparison of an image HiRISE of 2007 and an image of 1999 recorded by the Mars Global Surveyor of the dunes in the Nili Patera caldera, measuring therefore the motion of the dunes on a longer period of time than that previously investigated.

**Keywords**: Edge detection, Sobel filter, GIMP, Image processing, Google Mars, Dune motion, Satellite images, HiRISE, Mars Reconnaissance Orbiter, Mars Global Surveyor.

**1. Introduction**
Since the edge detection approach is often involved in the digital image processing, we have several methods that we can apply to satellite images to detect, for instance, the boundaries of some landforms and their evolution. Classical procedures are based on the filtering with gradients and on operators such as those of Roberts, Previt and Sobel [1]. More sophisticated procedures are based on the operators of Hueckel and Humel, and on the estimation of statistical characteristics of the image as in the LoG and Canny detectors [1]. Besides, these procedures, several new methods have been proposed on grey-scale and color images; among them for instance, those proposed in [1-4]. Some edge detection filters are also available in the GNU image manipulation program (GIMP). They are the edge, Laplace, Neon, Sobel and the difference of Gaussians, the details of which are given in [5]. Here, we will apply the GIMP Sobel filter as an example of the use of an edge detection method in enhancing the visibility of the motion of a dune on Mars.

**2. Moving dunes on Mars**
Only recently the dunes on Mars have been recognized as moving objects. This result has been achieved by means of the images coming from the satellites orbiting on Mars. These images showed that the dunes in the Gale Crater for instance, move at a rate of about 0.4 m/yr [6]. To investigate this motion the researchers used some images from the High Resolution Imaging Science Experiment (HiRISE), the camera onboard the Mars Reconnaissance Orbiter [7]. The researches evidenced also the sand motion in the Herschel Crater and in the Nili Patera caldera on Mars [8,9,10]. In [10], it had been concluded that the rates of landscape modification on Mars and Earth are similar. An example of two HiRISE images, recorded three years apart, are given at Ref.11, showing the motion of a dune of the Nili Patera dune field. However, images of the same dune field are available, which had been recorded several years before by the camera of the Mars Global Surveyor.
At Ref.12, we can find the HiRISE image (on the left in Figure 1). It is showing a set of dark sand dunes within the northeastern edge of the dune field in Nili Patera, which is a volcanic caldera in the region of the Syrtis Major on Mars. The dunes have a crescent shape, and therefore are defined as "barchans" in analogy with the dunes on Earth. The acquisition date of the image was 13 October 2007 (for more details, see Ref.12). At Ref.13, we can find an image with a part of the same dune field recorded by the Mars Orbiter Camera on the Mars Global Surveyor spacecraft on March 11, 1999 (portion of Mars Orbiter Camera image FHA-00451, NASA/JPL/Malin Space Science, reproduced on the right of the Figure 1). The time interval between the HiRISE and the MOC images is of about 8 years and therefore we expect a larger displacement of the dunes, a respect to that shown in Ref.11, using two images, recorded three years apart. However, it is necessary to remark that HiRISE and MOC images have different resolutions and slightly different orientations.

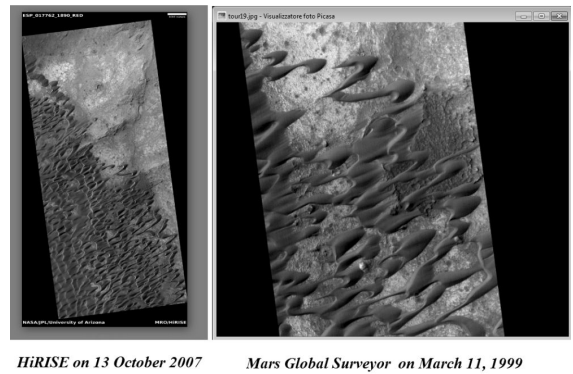

**Fig.1 – The images of the Nili Patera dune field that we can use for the study of moving dunes on Mars.**

## 3. Using GIMP and Sobel filtering

Let us try to compare the images using GIMP in order to find the motion of the dunes. We can select some dunes and adjust brightness and contrast [14], such as we did in some recent papers [15,16] to see the moving dunes on Earth. In the following Figure 2, a barchans of the Nili Patera caldera field is shown: on the left, the MOC image of 1999 and on the right the HiRISE image of 2007. We can see the displacement of the toe of the dune: both images have been processed with GIMP to adjust brightness and contrast, to enhance the visibility of the surface of the ground. However, the images shown in Figure 2 can be further processed with the Sobel filter, adjusting its threshold to low levels. Here the filtered images are shown in inverted colors (Figure 3). The image in inverted colors can be used a layer and added to the corresponding original image of Figure 2. The result is proposed in the Figure 4.

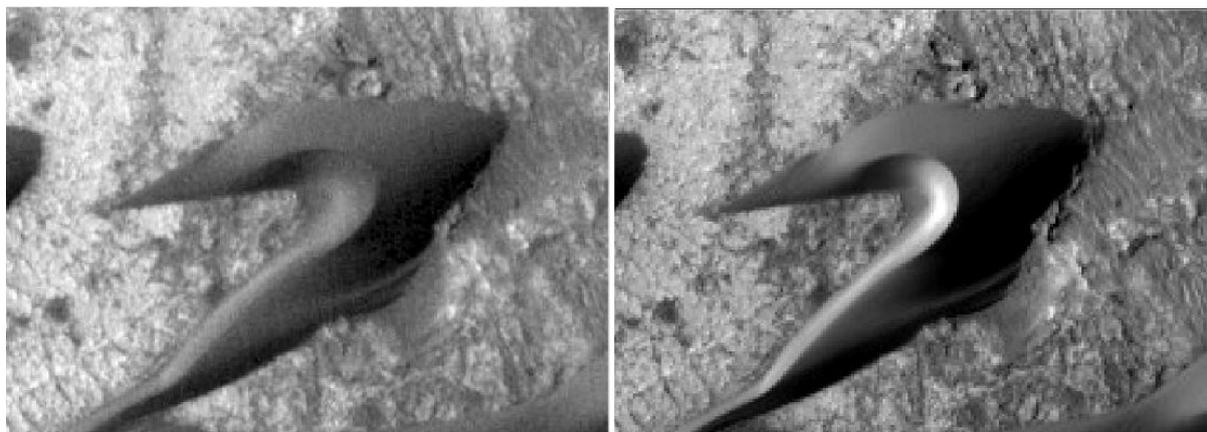

**Fig.2 – A barchan of the Nili Patera dune field in the Mars Orbiter Camera image (1999) on the left and in HiRISE on the right (2007). Note the displacement of the toe of the dune. Both images have been processed with GIMP to adjust brightness and contrast, to enhance the visibility of the ground on which the dune is moving and have then some reference points on it.**

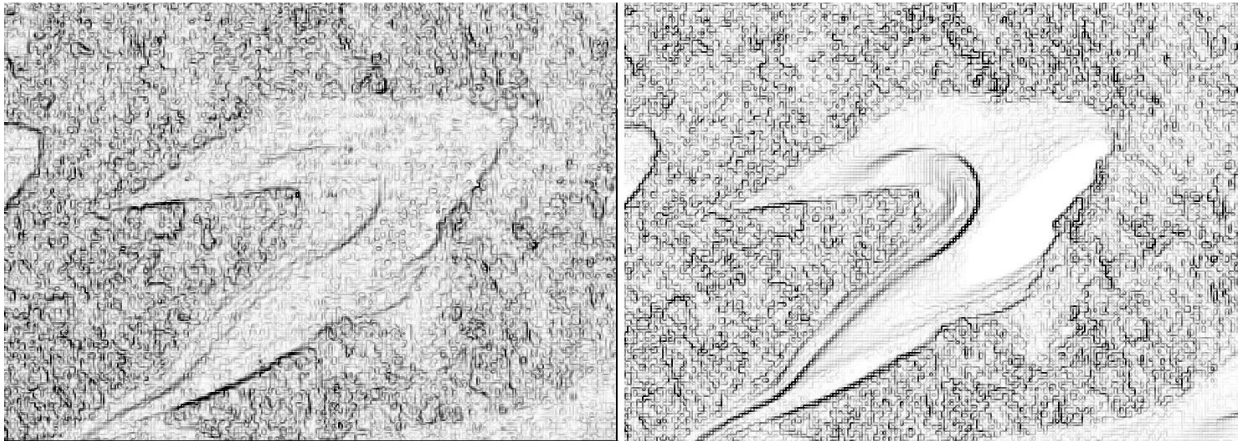

**Fig.3** – The images in the Figure 2 have been processed with the Sobel filter of GIMP, adjusting the threshold. Here the filtered images are shown in inverted colors. These two images can be used as layers and added to the original images of Figure 2. The result is proposed in the Figure 3.

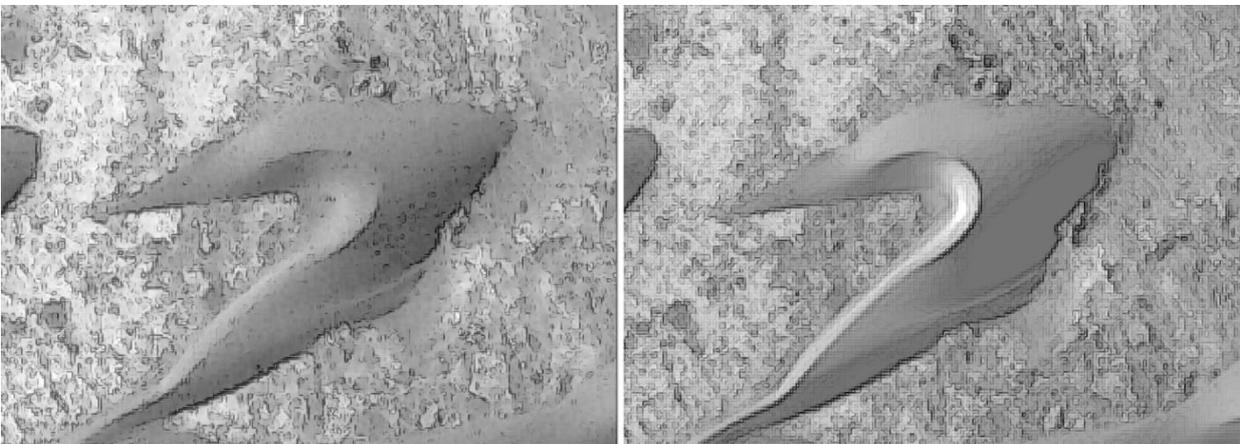

**Fig.4** – The images in Figures 2 and 3 are used as layers and merged together in a new image. With respect to the images in Figure 2, we can see more clearly the relative displacement of the dune on the ground of the crater.

As a result of the Sobel processing and the use of its output as a layer added to the original image, we can see more clearly the displacement of the dune with respect to the ground. In [14] we proposed an estimation of the speed of the toe of this dune, which is approximately of 10 meters in eight years. This seems a large value, however, this is one of the smaller dunes in this dune field. On Earth we know that small dunes move faster that the large dunes: if we consider, as did in [10], the same behavior for the dunes on Mars, this result is well posed.

### 4. Conclusion
Here we have discussed a simple image processing which is using an edge detection filter to compare a HiRISE image and a Mars Global Surveyor image, and investigate the motion of a barchan in the Nili Patera caldera on Mars. As the use of the GIMP processing is showing, an image processing and Sobel filtering allow a better comparison of images coming from two collections, in this case those of the Mars Global Surveyor and HiRISE.

**References**
[1] B. Marinov, V. Gospodinova, Adaptive procedure for boundary detection in satellite images, 3rd International Conference on Cartography and GIS, 15-20 June, 2010, Nessebar, Bulgaria, 11 pages.

[2] F.A. Pellegrino, W. Vanzella, V. Torre, Edge Detection Revisited, IEEE Transactions on Systems, Man, and Cybernetics — Part B: Cybernetics, volume 34, issue 3, June 2004, pages 1500-1518.

[3] T. Shimada, F. Sakaida, H. Kawamura, T. Okumura, Application of an edge detection method to satellite images for distinguishing sea surface temperature fronts near the Japanese coast, Remote Sensing of Environment, volume 98, issue 1, 30 September 2005, pages 21–34.

[4] A.C. Sparavigna, Color Dipole Moments for Edge Detection. The Computing Research Repository (CoRR), April 2009, 8 pages, oai:arXiv.org:0904.0962

[5] http://docs.gimp.org/en/filters-edge.html

[6] S. Silvestro, D.A. Vaz, R.C. Ewing, A.P. Rossi, L.K. Fenton, T.I. Michaels, J. Flahaut, and P.E. Geissler, Pervasive aeolian activity along rover Curiosity's traverse in Gale Crater, Mars, Geology, volume 41, issue 4, 2013, pages 483-486.

[7] P.E. Geissler, M.E. Banks, N.T. Bridges, and S. Silvestro, HiRISE Observations of Sand Dune Motion on Mars: Emerging Global Trends, Third International Planetary Dunes Workshop: Remote Sensing and Data Analysis of Planetary Dunes, held June 12-15, 2012 in Flagstaff, Arizona. LPI Contribution No. 1673, pages 44-45, 06/2012.

[8] M. Cardinale, S. Silvestro, D.A. Vaz, T.I. Michaels, L. Marinangeli, G. Komatsu and C.H. Okubo, Evidences for Sand Motion in Herschel Crater (Mars), 44th Lunar and Planetary Science Conference, held March 18-22, 2013 in The Woodlands, Texas. LPI Contribution No. 1719, page 2259, 03/2013.

[9] S. Silvestro, L.K. Fenton, D.A. Vaz, N. Bridges and G.G. Ori, Ripple Migration on Active Dark Dunes in Nili Patera (Mars), Second International Planetary Dunes Workshop: Planetary Analogs - Integrating Models, Remote Sensing, and Field Data, held May 18-21, 2010 in Alamosa, Colorado. LPI Contribution No. 1552, pages 65-66, 05/2010.

[10] N. T. Bridges, F. Ayoub, J-P. Avouac, S. Leprince, A. Lucas and S. Mattson, Earth-like sand fluxes on Mars, Nature, volume 485, pages 339–342, 17 May 2012.

[11] NASA, http://www.nasa.gov/mission_pages/MRO/multimedia/pia15295.html

[12] N. Bridges, Sand Dunes in Nili Patera Caldera, PSP_005684_1890 Science Theme: Volcanic Processes, hirise.lpl.arizona.edu/ESP_017762_1890

[13] Lunar and Planetary Institute, A spacecraft tour of the solar system, www.lpi.usra.edu/publications/slidesets/ss_tour/slide_19.html

[14] A.C. Sparavigna, Moving sand dunes on Mars, 23/08/2013, available at PORTO, http://porto.polito.it/2513080/

[15] A.C. Sparavigna, Moving dunes on the Google Earth, arXiv:1301.1290 [physics.geo-ph], arXiv, 4 January 2013.

[16] A.C. Sparavigna, A Study of Moving Sand Dunes by Means of Satellite Images, The International Journal of Sciences, volume 2, issue August, 2013, pages 33-42.